\documentclass[journal]{IEEEtran}
\usepackage{cite}
\usepackage{multirow}
\usepackage{amsmath}
\usepackage{graphicx}
\usepackage{booktabs}
\usepackage{subfigure}
\usepackage{stfloats}
\usepackage{makecell}

\usepackage{color}
\usepackage{amssymb}

\ifCLASSINFOpdf
\else
\fi

\begin{document}
%
\title{Spatial-Spectral Residual Network for Hyperspectral Image Super-Resolution}
%
%
%

\author{Qi~Wang,~\IEEEmembership{Senior Member,~IEEE,}
        Qiang~Li,~
        and~Xuelong~Li,~\IEEEmembership{Fellow,~IEEE}
\thanks{ The authors are with the School of Computer Science and the Center for OPTical IMagery Analysis and Learning (OPTIMAL), Northwestern Polytechnical University, Xi'an 710072, China (e-mail: crabwq@gmail.com, liqmges@gmail.com, xuelong\_li@nwpu.edu.cn) (\textit{Corresponding author: Xuelong Li.})}}

\maketitle

\begin{abstract}
Deep learning-based hyperspectral image super-resolution (SR) methods have achieved great success recently. However, most existing models can not effectively  explore spatial information and spectral information between bands simultaneously,  obtaining relatively low performance. To address this issue, in this paper, we propose a novel spectral-spatial residual network for hyperspectral image super-resolution (SSRNet). Our method can  effectively explore  spatial-spectral information by using 3D convolution instead of 2D convolution, which  enables  the network to  better extract potential information.  Furthermore, we design  a spectral-spatial residual module (SSRM) to adaptively learn more effective features  from all the hierarchical features in units through local feature fusion, significantly improving the performance of the algorithm. In each unit, we employ spatial and temporal separable 3D convolution to extract spatial  and spectral information, which not only reduces unaffordable memory usage and high computational cost, but also makes the network easier to train. Extensive evaluations and comparisons on three benchmark  datasets demonstrate that the proposed approach achieves superior performance in comparison to existing state-of-the-art methods.
\end{abstract}


\IEEEpeerreviewmaketitle

\section{Introduction}
\IEEEPARstart{H}{yperspectal} imaging system collects surface information in tens to hundreds of continuous spectral bands to acquire hyperspectral image. Compared with multispectral image or natural image, hyperspectral image has more abundant spectral information of ground objects, which can reflect the subtle spectral properties of the measured objects in detail \cite{li2019efficient}. As a result, it is widely used in various fields,  such as mineral exploration \cite{sabins1999remote}, medical diagnosis \cite{Lin2018Dual}, plant detection \cite{lowe2017hyperspectral}, etc.  However, the obtained hyperspectral image is often low-resolution because of the interference of environment and other factors, which limits the performance of high-level tasks, including   change detection \cite{Wang2019GETNET}, image classification \cite{Wang2019Locality}, etc.

\begin{figure}[t]
	\centering
	\includegraphics[height=6cm,width=0.48\textwidth]{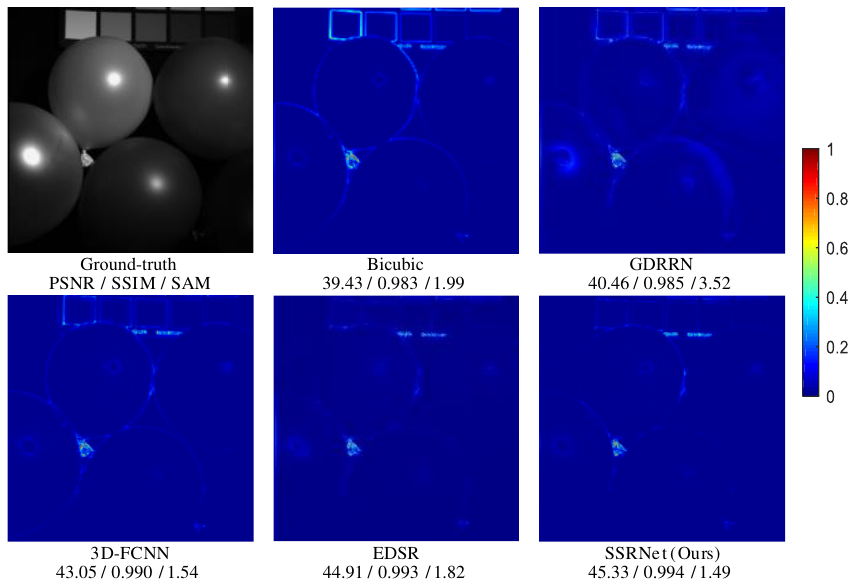}
	\caption{Comparisons of our SSRNet  with existing methods on hyperspectral image SR for scale factor $\times$4. The absolute error map of one band is showed between reconstructed hyperspectral image and ground-truth. In general, the bluer the absolute error map is, the better the restored image is. }
	\label{fig:fig1}
\end{figure}

To  better and accurately  describe  the ground objects, the hyperspectral image super-resolution (SR)  is proposed \cite{Xie2019Hyperspectral, Dong2016Hyperspectral, akgun2005super}. It aims to restore high-resolution hyperspectral image from degraded low-resolution hyperspectral image. In practical application, the objects in the image are  often detected or recognized according to the spectral reflectance of the object. Therefore, spectral and spatial resolution should be considered simultaneously for hyperspectral image SR, which is different from  natural image SR in computer vision \cite{hu2019channel}.

 Since the spatial resolution of hyperspectral image is lower than that of RGB image \cite{Ying2018Unsupervised}, existing  methods  mainly  fuse high-resolution RGB image with  low-resolution hyperspectral image \cite{Lim2015RGB, Akhtar2016Hierarchical, Akhtar2015Bayesian}. For instance,  Kwon \textit{et al.} \cite{kwon2015rgb} utilize the RGB image corresponding to high-resolution hyperspectral image to obtain poorly reconstructed image. Then the image in local is refined by sparse coding to obtain better SR image. Under the prior knowledge on  spectral and spatial transform responses, Wycoff \textit{et al.} \cite{Wycoff2013A} formulate the SR problem into non-negtive sparse factorization.  The problem is effectively addressed by alternating direction method of multipliers \cite{boyd2011distributed}.  These methods realize hyperspectral image SR under the guidance of RGB images generated by the same camera spectral response (CSR)\footnote{http://www.maxmax.com/aXRayIRCameras.htm}, ignoring the differences of CSR between datasets or scenes. Suppose that the same CSR value is used in the process of reconstruction, which will obviously lead to the poor robustness of the algorithm. To address this issue, Fu \textit{et al.} \cite{Fu2019Hyperspectral} design the CSR function selection layer, which can automatically select the optimal CSR according to a particular scene. In addition to the CSR function selection mechanism, the method simulates CSR as the convolutional layer to learn the optimal CSR function, significantly improving the performance of hyperspectral image SR. However, such a scheme requires the pair of images to be well registered, which  is usually difficult to follow in practice. Moreover, the scholars claim that  these algorithms  are unsupervised, but they are not actually unsupervised in that the ground-truth for RGB image is adopted during reconstruction. 

The research of natural image SR has achieved great success in recent years due to the powerful representational ability of convolution neural networks (CNNs) \cite{Anwar2019A, zhang2018residual}. Its main principle is to learn the mapping function between low-resolution and high-resolution images in a supervised way. The typical methods include SRCNN \cite{dong2014learning}, EDSR \cite{Lim2017Enhanced}, and SRGAN \cite{ledig2017photo}, etc. Due to the satisfying performance in natural image SR, the scholars apply these methods for hyperspectral image SR.  Inspired by deep recursive residual network \cite{tai2017image},  Li \textit{et al.} \cite{Li2018Single} propose grouped deep recursive residual network (GDRRN) to execute hyperspectral image SR task in space. As  we mentioned earlier, obviously, this method does not take into account spectral resolution and thus may lead to spectral distortion of the restored hyperspectral image.  Considering this limitation, Mei \textit{et al.}  \cite{Mei2017Hyperspectral} present 3D full convolution neural network (3D-FCNN) to explore the relationship of the spatial information and adjacent pixels between spectra. Although this method effectively uncovers spatial information and spectral information between bands, it changes the size of the estimated hyperspectral image, which is not suitable for the purpose of image reconstruction.

To address these drawbacks, in this paper, we propose a novel spectral-spatial residual network for hyperspectral image super-resolution (SSRNet). Our method learns the mapping function in a supervised way without using RGB image corresponding to  high-resolution hyperspectral image. The whole network  uses 3D convolution  to extract hyperspectral image features instead of 2D convolution. In each spatial-spectral residual module (SSRM), the network can adaptively learn more effective spatial and spectral features from all the hierarchical units. To reduce unaffordable memory usage and high computational cost, we employ  separable 3D convolution to  extract spatial information and spectral information between bands in residual unit. Through three evaluation indexes, we demonstrate that the performance of  SSRNet  is superior to the state-of-the-art  hyperspectral image SR approaches based on deep learning   on three datasets. Besides, our proposed SSRNet  generates more realistic visual results compared with other methods, as shown in Fig. \ref{fig:fig1}. 

In summary, our main contributions are follows:

$\bullet$ A novel spatial-spectral residual network (SSRNet) is proposed to reconstruct hyperspectral image. The network can explore the spatial information and spectral information between bands without changing the size of hyperspectral image. It significantly enhances  the performance.

$\bullet$ The spatial-spectral residual module (SSRM) is designed to adaptively preserve the accumulated features through local feature fusion. It makes full use of all the hierarchical features in the unit, which enables the network to fully extract the features of hyperspectral images. 

$\bullet$ Spatial and temporal separable 3D convolution is employed to extract spatial and spectral features in each unit, respectively.  It can reduce unaffordable memory usage and high computational cost, and make the network easier to train.


The remainder of this paper is organized as follows: Section II describes existing hyperspectral image SR with CNNs and the detailed 3D convolution. Section III introduces our proposed SSRNet, including network structure, spectral-spatial residual module, skip connections, etc. Then,  experiments on benchmark datasets are performed to  verify our method in Section IV. Finally, Section V gives the conclusion. 

\section{Related Work}
There exists an extensive body of literatures on hyperspectral image SR. Here we first outline several  deep learning-based hyperspectral image SR methods. In order to better understand the proposed method, we then give a brief introduction to 3D convolution.

\subsection{Hyperspectral Image SR with CNNs}
\begin{figure}[t]
	\centering
	\includegraphics[height=2cm,width=0.4\textwidth]{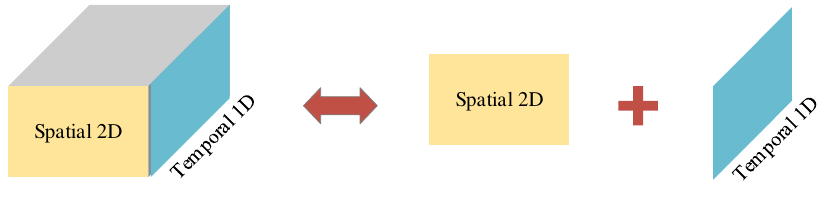}
	\caption{Spatial and temporal separable 3D convolution.}
	\label{fig:threeD}
\end{figure}
Recently,  deep learning-based methods \cite{han2018ssf} have achieved remarkable advantages in the field of hyperspectral image SR. Here, we will briefly introduce several methods with CNNs.  Li \textit{et al.} \cite{Li2017Hyperspectral} propose  a deep spectral difference convolutional neural network (SDCNN) by using five convolutional layers to improve spatial resolution. Under spatial constraint strategy, it makes the reconstructed hyperspectral image  preserve spectral information through post-processing.  Jia \textit{et al.} \cite{Jia2018Hyperspectral} present spectral-spatial network (SSN), including spatial and spectral sections. They try to  learn the mapping function between low-resolution and high-resolution images  and fine-tune spectrum.  Yuan \textit{et al.} \cite{Yuan2017Hyperspectral} utilize  the knowledge from natural image to  restore high-resolution hyperspectral image by transfer learning, and collaborative nonnegative matrix  factorization is proposed to enforce collaborations between low-resolution and high-resolution hyperspectral images. All of these methods need two steps to achieve image reconstruction, that is, the algorithm first improves the spatial resolution. To avoid spectral distortion, some constraint criteria are then employed to retain the spectral information.  It is clear that the spatial resolution may be changed while maintaining the spectral information. 

Considering this issue, Li \textit{et al.} \cite{Li2018Single} and Wang \textit{et al.} \cite{ wang2017deep} introduce spectral angle error and set a new loss function by combining it with the mean square error. When training the network, these methods combine two error functions and deliberately reduce the distortion of the spectrum. However, it  affects the performance of the reconstructed spatial resolution. Unlike natural image, the hyperspectral image has  tens to hundreds of continuous spectral bands. Mei \textit{et al.} \cite{Mei2017Hyperspectral} take advantage of this property of hyperspectral image and adopt 3D convolution to extract the features, which effectively retains the information of the original spectrum and improves the performance of image SR. However, the size of reconstructed image is changed.

\subsection{3D Convolution}
For natural image SR, the scholars usually employ 2D convolution to extract the features and obtain good performance \cite{li2019fast, dai2019second}. As we introduced earlier, the hyperspectral image contains  many continuous bands, which results in  a significant characteristic that  there is a great correlation between adjacent bands \cite{Wang2019Hype}. If we directly utilize 2D convolution to conduct  hyperspectral image SR task, it will make it impossible to effectively exploit potential features between bands. Therefore, in order to make full use of this characteristic, we design  network by using 3D convolution to analyze the spatial and spectral features of hyperspectral image in our paper.

Since 3D convolution takes into account the inter-frame motion information in the time dimension, it is widely used in video classification \cite{tran2019video}, action recognition \cite{zhou2018mict} and other fields. Unlike 2D convolution, the 3D convolution operation is implemented by convolving a 3D kernel with  feature maps. Intuitively, the number of parameters of the training network using 3D convolution is an order of magnitude more than that of the  2D convolution. To address this problem, Xie \textit{et al.} \cite{xie2018rethinking} develop  typical separable 3D CNNs (S3D) model to accelerate  video classification. In this model, the standard 3D convolution is replaced by spatial and temporal separable 3D convolution (see Fig. \ref{fig:threeD}), which demonstrates that this way  can effectively reduce the number of parameters while still maintain good performance.

\section{Proposed Method}
\begin{figure*}[t]
	\centering
	\includegraphics[height=3.6cm,width=1.0\textwidth]{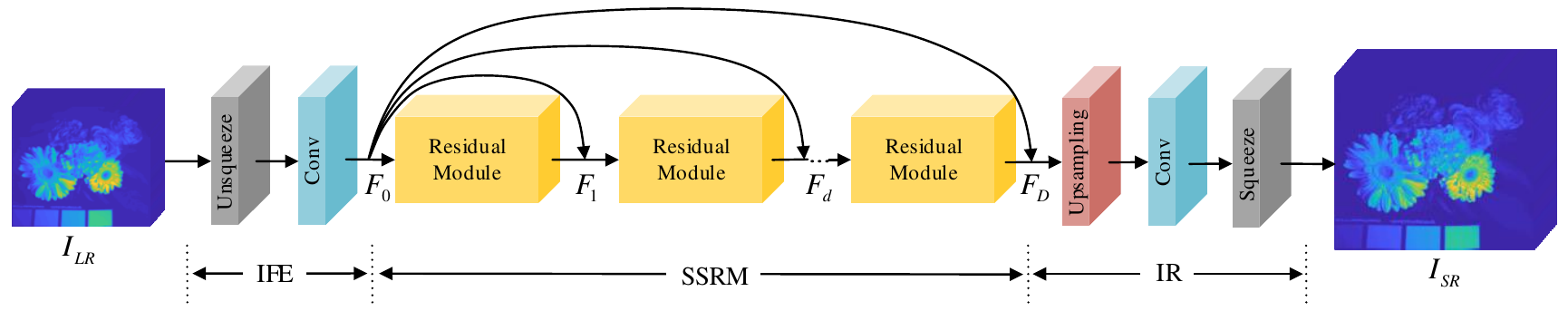}
	\caption{Overall architecture of our proposed SSRNet. }
	\label{fig:flowchart}
\end{figure*}

\subsection{Network Structure}
In this section, we will detail overall architecture of our SSRNet, whose flowchart is shown in Fig. \ref{fig:flowchart}. As can be seen from this figure, our method mainly consists of three parts: initial feature extraction (IFE) subnetwork, spatial-spectral residual module (SSRM) subnetwork, and image reconstruction (IR) subnetwork.  Let $I_{LR}\in R^{ W \times H \times L}$ and $I_{SR}$ represent the input low-resolution hyperspectral image and the output reconstructed hyperspectral image, where  $W$ and $H$ are the width and height of each band, and $L$ represents  the total number of the bands in hyperspectral image. In order to employ 3D convolution, we need unsqueeze $I_{LR}$ into four dimensions ($ W \times H \times L \times 1 $) at the beginning of the network. Then,  a standard 3D convolution is applied to extract shallow features about $I_{LR}$, i.e.,
\begin{equation}
	F_0 = f_{c}(Unsqueeze(I_{LR})),
\end{equation}
where $Unsqueeze(.)$ means the input hyperspectral image is expanded four dimensions, and $f_{c}$ denotes 3D convolution operation. The initial features of $F_0$ is fed into spatial-spectral residual module, which is described in detail in Section \ref{section 3.b}. After $D$ residual modules and global skip connection, the deep feature maps $F_D$ are denoted as 
\begin{equation}
	F_D=F_0+M_D(M_{D-1}(...M_1(F_0)+F_0...)+ F_0),
\end{equation}
where $M_d(.)$ denotes the operation of the  $d$-th residual module. With respect to the impact of the number of residual module $D$ in our network, we will analyze it in Section \ref{ND}. For IR sub-network, we use transposed convolution layer to upsample these feature maps to the desired scale via scale factor $r$, which is followed by  a convolution layer.  After squeeze process, the output size  becomes $W \times H \times L$.  Finally, the output of SSRNet can  be  obtained by 

\begin{equation}
	I_{SR} = squeeze(f_{c}(f_{up}(F_D))),
\end{equation}
where $f_{up}$ and $squeeze(.)$ are the functions for upsampling and squeeze, respectively.


\subsection{Spatial-Spectral Residual Module}
\begin{figure*}[t]
	\centering
    \includegraphics[width=0.95\textwidth]{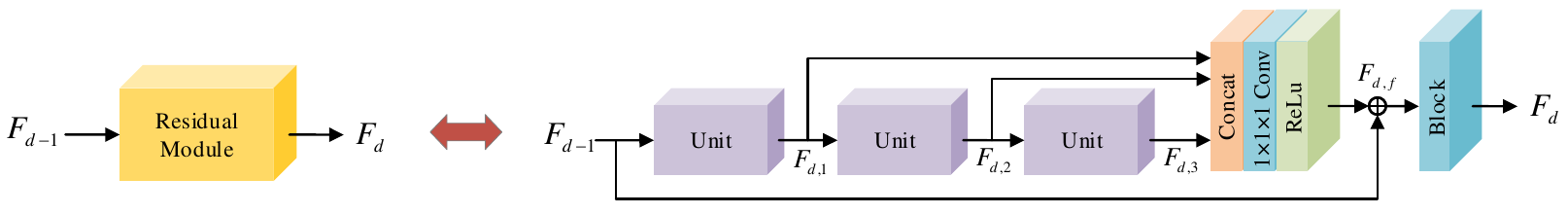}
    \caption{Architecture of  the $d$-th spectral-spatial residual module (SSRM). The module contains three units, local feature fusion, and a block. The feature maps from $F_{d-1}$ are first fed into the first unit. After two units, the output of each unit is  concatenated together to fuse these features of different depths. More effective features are attached to block after local residual learning and  the output of the module $F_d$ is finally obtained.}
	\label{fig:ssrm}
\end{figure*}
\label{section 3.b}
The  architecture of spatial-spectral residual module (SSRM) is  illustrated in Fig. \ref{fig:ssrm}. As  provided in this figure, the module mainly contains three residual units, local feature fusion, and a block. In the $d$-th SSRM, suppose $F_{d-1}$ and $F_{d}$  are the input and output feature maps, respectively. Under the local residual connection,  the output $F_{d}$ of the $d$-th SSRM can be defined as 
\begin{equation}
	F_d = f_B(F_{d,f} + F_{d-1}), 
\end{equation}
where $f_B$ is the function of the block. Next, we will present the details about the proposed residual unit and block.

\subsubsection{Residual Unit}
\begin{figure}[t]
	\centering
	\subfigure[Separable 3D convolution]{
		\label{fig:block(a)}
		\includegraphics[width=0.39\textwidth]{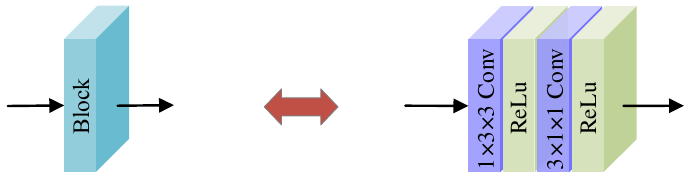}}
	\subfigure[Residual unit]{
		\label{fig:block(b)}
		\includegraphics[width=0.48\textwidth]{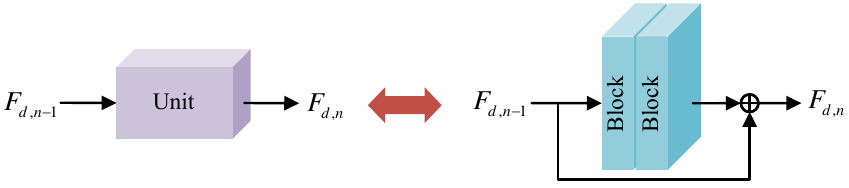}}				
	\caption{Architecture of the $n$-th residual unit.}
	\label{fig:block}
\end{figure}

As we said in Section II, the previous work use  spatial and temporal separable 3D convolution to represent the standard 3D convolution for video classification, i.e, the size of the filter  $k \times k \times k$ is modified as $k \times 1 \times 1$ and $1 \times k \times k$,  which has been proven to perform better. To reduce  unaffordable memory usage and high computational cost, in our paper, we use this method to replace the standard 3D convolution in the block.  Specifically, the filter $k \times 1 \times 1$ is used  to extract the features between spectra, and the filter $1 \times k \times k$ is adopted to  extract the spatial features of each band. Moreover,  we add the rectified linear unit (ReLU) after each convolution operation (see Fig. \ref{fig:block(a)}). Finally,  the block can be formulated as 
\begin{equation}
	f_{B}(.)= \sigma(f_{c}(\sigma(f_{c}(.)))),
\end{equation}
where $\sigma$ denotes the ReLU activation function. In terms of this way, it can not only effectively mine the potential information between spectra, but also speed up the implementation of the algorithm. Since the size of convolution kernel $3 \times 3$ can extract image features well for natural image SR, in our work, the parameter $k$ of convolution is set to 3.

Now we present the proposed residual unit, which is shown in Fig. \ref{fig:block(b)}. Let $F_{d,n-1}$ and $F_{d,n}$ are the input and output feature maps of  the $n$-th unit in the $d$-th SSRM, respectively. Through the local skip connection and two blocks, the  output feature maps $F_{d,n}$ can be obtained by 
\begin{equation}
	F_{d,n}=f_B(f_B(F_{d,n-1}))+F_{d,n-1}.
\end{equation}
By doing so, it can not only greatly reduce the computational cost, but also simultaneously learn spectral and spatial information of hyperspectral image.

\subsubsection{Local Feature Fusion}
To make the network learn more useful information, we design  local feature fusion strategy (see Fig \ref{fig:ssrm}) to adaptively retain the cumulative features, which enables the network can fully extract hyperspectral image features. Specifically,  the features from different units are first concatenated to learn fusion information. In order to do a local residual learning between the fused result and the input $F_{d-1}$, it is necessary to reduce the number of features. Thus, we add a convolution layer with the size $1\times 1 \times 1$  after concatenation to adaptively retain valid information. Besides, we also set the ReLU activation function after convolution. As a result, the output of local feature fusion $F_{d,f}$ is formulated as 
\begin{equation}	
F_{d,f} = \sigma (f_{c}(Concat(F_{d,1}, F_{d,2}, F_{d,3}))),
\end{equation}
where $Concat(.)$ denotes concatenation function of different hierarchical features.

\subsection{Skip Connections}

As the depth of the network increases, the weakening of information flow and the disappearance of gradient hinder the training of the network. Recently, there are many ways to solve these problems. For instance, He \textit{et al.} \cite{he2016deep} first utilize skip connection between layers so as to improve the information flow and make it easier to train. To fully explore the advantages of skip connection, Huang \textit{et al.} \cite{Huang2017D} propose DenseNet. The network has the advantages of strengthening feature propagation, supporting feature reuse, and reducing the number of parameters. 

For SR task, the input low-resolution  image is greatly similar to the output high-resolution  image, that is, the low-frequency information carried by the low-resolution  image is similar to that of the high-resolution  image \cite{kim2016accurate}.  According to this characteristic, the researchers use dense connections to enhance the information flow  of the whole network and  alleviate the  disappearance of the gradient  for natural image SR, thus effectively improving the performance of the algorithm. Therefore, we add several global residual connections in our network. Since the shallow network can retain more  edge or texture information of hyperspectral image, the feature maps from IFE are fed into the the back  of each module, which can enhance the performance of the entire network.

\subsection{Network Learning}
For network training,  the SSRNet  is optimized by minimizing the difference between reconstructed hyperspectral image $I_{SR}$ and corresponding ground-truth hyperspectral image $I_{HR}$. 
Mean square error (MSE) is often used as loss function to study the parameters of the network for hyperspectral image SR algorithms based on deep learning \cite{Li2017Hyperspectral}. Additionally, some methods design two terms in loss function to minimize the difference, including MSE and spectral angle mapping (SAM) \cite{wang2017deep, Li2018Single}. In fact, these loss functions do not make the network converge better and obtain poor results, which is proved in the experiment section. For natural image SR, as far as we know, many networks in recent years usually use $L1$  as  loss function, and the experiments also demonstrate that the $L1$ can obtain more powerful performance and convergence \cite{Anwar2019A}. Therefore, in this paper, we refer to the natural image SR method and adopt $L1$ as the loss function of our designed network.  The loss function of SSRNet is
\begin{equation}
	{\cal L}(I_{SR},I_{HR};\theta) = \frac{1}{M}\sum\limits_{m = 1}^M {||{I_{HR}}^{(m)} - {I_{SR}}^{(m)}|{|_1}},
\end{equation}
where $M$ is the number of training patches and $\theta$ denotes the parameter set of the SSRNet network.
\section{Experiment}
To verify the effectiveness of the  proposed SSRNet, in this section, we first introduce three public datasets, implementation details, and evaluation indexes. We then analyze the proposed method from many aspects, including loss function analysis, ablation study, etc. Finally, we assess  the performance of our SSRNet by comparisons to the state-of-the-art methods.

\subsection{Datasets}
\begin{figure}[t]
	\centering
	\includegraphics[height=5cm,width=0.48\textwidth]{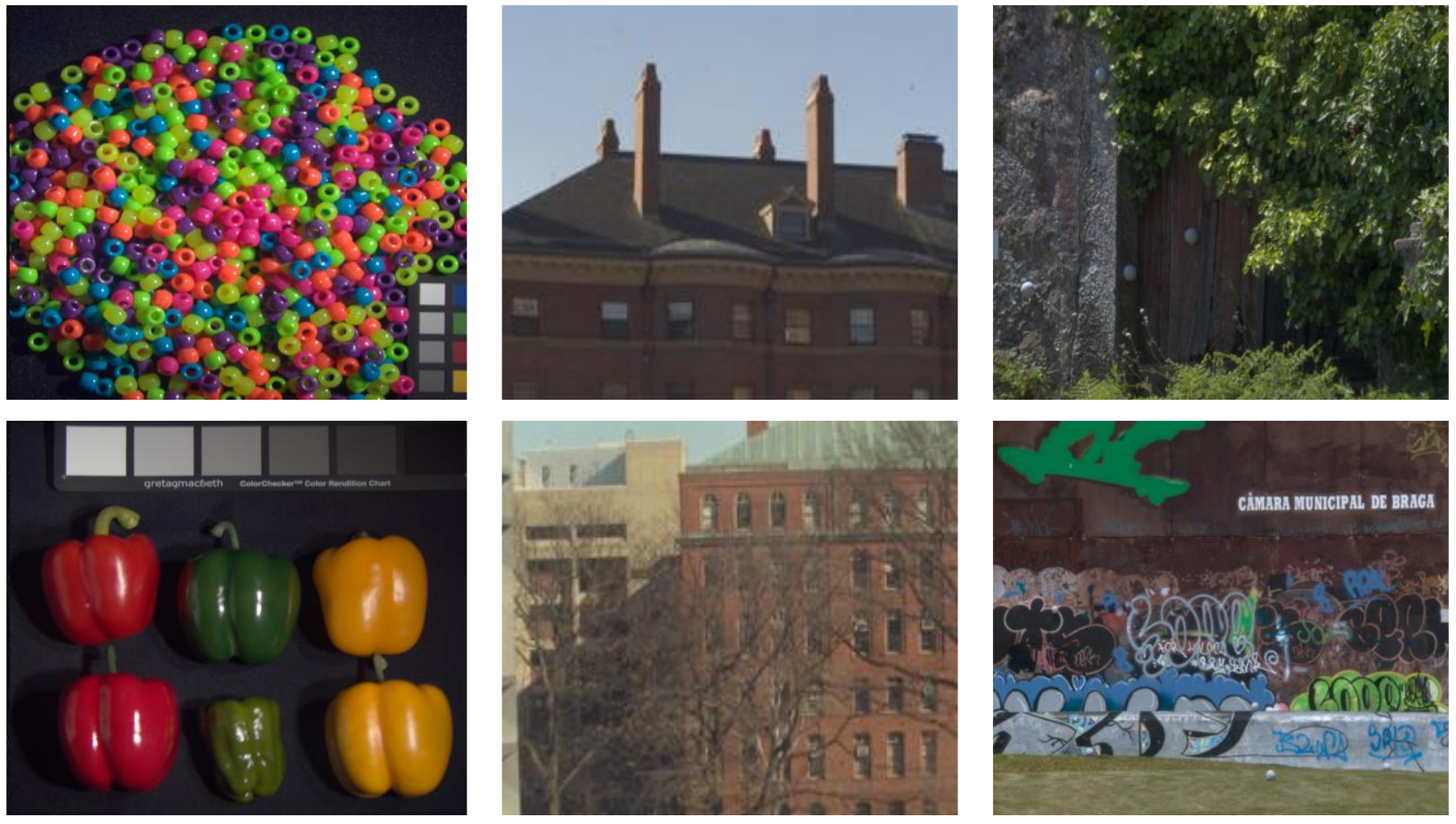}
	\caption{Some RGB images corresponding to hyperspectral images on three datasets.}
	\label{fig:data}
\end{figure}
\subsubsection{CAVE} The CAVE dataset\footnote{http://www1.cs.columbia.edu/CAVE/databases/multispectral/}  is gathered by cooled CCD camera at a 10nm step from 400 nm to 700 nm (31 bands) \cite{yasuma2010generalized}. The dataset contains 31 scenes, divided into 5 sections: real and fake, skin and hair, paints, food and drinks, and stuff. The size of all hyperspectral image is $512 \times 512 \times 31$ in this dataset. Each band is stored as a 16-bit grayscale PNG image. 
\subsubsection{Harvard} The Harvard dataset\footnote{http://vision.seas.harvard.edu/hyperspec/explore.html} is obtained by Nuance FX, CRI Inc. camera   in the wavelength range of 400 nm to 700 nm. \cite{chakrabarti2011statistics}. The dataset consists of  77 hyperspectral images of real-world indoor or outdoor scenes under daylight illumination. The size of each hyperspectral image is $1040 \times 1392 \times 31$ in this dataset. Unlike CAVE dataset, this dataset is stored as .mat file.  

\subsubsection{Foster} The Foster dataset\footnote{https://personalpages.manchester.ac.uk/staff/david.foster/Local\_Illuminati-

on\_HSIs/Local\_Illumination\_HSIs\_2015.html} is collected using a low-noise
Peltier-cooled digital  camera (Hamamatsu, model C4742-95-12ER) \cite{nascimento2016spatial}. The dataset includes 30 images from the Minho region of Portugal during late spring and summer of 2002 and 2003. Each hyperspectral image has 33 bands  with the size of 1204 $\times$ 1344 pixels. Similarly, the dataset is also stored as .mat file.  Some RGB images corresponding to hyperspectral images are shown in Fig. \ref{fig:data}.


\subsection{Implementation Details}
As mentioned earlier, different datasets are gathered by different hyperspectral cameras, so we need to train and test each dataset individually, which is different from the natural image SR. In our work, 80\% of the samples are randomly selected as training set, and the rest are used for testing. 

For the training phase, since there are too few images in these datasets for deep learning algorithm, we augment the training data by randomly selecting 24 patches with the size of $32 \times 32 \times L$. Each patch is  horizonta flipped, rotated ($90^{\circ}$, $180^{\circ}$, and $270^{\circ}$), and scaled (1, 0.75, and 0.5). According to scale factor $r$, these patches are downsampled as low-resolution hyperspectral images by  bicubic interpolation. Before feeding the mini-batch  into our network, we subtract the average value of the entire training images for patches. In our work, we set the size of filter as $3 \times 1 \times 1$ and $1 \times 3 \times 3$ in each convolution layer expect those for initial feature extraction and image reconstruction (the size of filter is set to $3 \times 3 \times 3$), and the number of filter for all layer in our network is set to 64. We initialize each convolutional filter using\cite{Yu2018Wide}. The ADAM optimizer with ${\beta _1 = 0.9, \beta _2 = 0.999}$ is employed to train our network.  The learning rate is initialized as $10^{-4}$ for all layers, which decreases by a half at every 35 epochs.

For the test phase, in order to improve the efficiency of the test, we only use the top left $512 \times 512$ region of  each test image for evaluation. Our method is conducted using the PyTorch framework with NVIDIA GeForce GTX 1080 GPU. 

\subsection{Evaluation Metrics}
To qualitatively measure the proposed SSRNet, three evaluation methods are employed to verify the effectiveness of the algorithm, including  peak signal-to-noise ratio (PSNR), structural similarity (SSIM), and spectral angle mapping (SAM). In general, the larger the PSNR and SSIM is and the smaller the SAM is, the better the performance of the reconstructed hyperspectral image is. 

\subsection{Model Analysis}
In this section, to verify the effectiveness of our proposed method, we  conduct sufficient experiments  from the following four aspects.
\subsubsection{Loss Function Analysis}
\begin{table}[tp]
	\centering
	\renewcommand\arraystretch{1.2}
	\caption{Loss function analysis  for scale factor $\times$2 on CAVE dataset.}
	\label{table:lossfun}
    \begin{tabular}{c|ccc}
	\toprule[0.8pt]	
	Loss Function & PSNR & SSIM & SAM  \\ \hline
	MSE   & 44.62     & 0.973   & 2.33    \\
	L1   & 44.99   & 0.974  & 2.23 \\
	0.5*MSE+0.5*SAM   & 43.22   & 0.970    & 2.46  \\ \toprule[0.8pt]	
   \end{tabular}
\end{table}

To demonstrate the effect of different loss functions,  the loss functions of \cite{wang2017deep}, \cite{Li2017Hyperspectral}, and L1 in our work are employed to train SSRNet on  CAVE dataset. The evaluation results are shown in Table \ref{table:lossfun}.  When adding SAM in loss function, it is clear that the spatial resolution has changed,  and the spectral distortion has become more serious. Moreover, the loss function containing  MSE and SAM gets  a lower PSNR value, which is mainly due to the fact that the loss function weakens the performance of spatial resolution. As seen from this table, L1 in our paper can achieve the best performance than other loss functions for  three indexes.  It verifies our method can effectively optimize the difference between  $I_{SR}$ and $I_{HR}$ using L1.

\subsubsection{Ablation Study}
Table \ref{table:ablation_study} shows the ablation study  on the impacts of local feature fusion (LFF) in module and global residual learning (GRL). We set the different combinations of components to analyze the performance of the proposed SSRNet. To simply  do  fair comparison,  our network with 3 modules is adopted to implement ablation investigation for scale factor $\times$2 on CAVE dataset.

First, without the local feature fusion and global residual learning (LFF0GRL0), the network yields  the worst performance. It mainly lacks of adequate learning of effective features, which also shows that spectral and spatial features can not be extracted  well without these components. Thus, these components are required in our network. Then, we add one of these components, LFF, to the baseline (LFF0GRL0). The performance of the network  is improved in PSNR and SAM. Accordingly, only GRL (denote as LFF0GRL1) is  added to the baseline.  Evaluation indexes attain relatively better than  the results of LFF0GRL0, except for SSIM. In short,  the experiments demonstrate that each component can clearly enhance the performance of the network. This indicates that each component plays a key role in  making the network easier to train. Finally, two components (LFF1GRL1) are attached to the baseline. The table exhibits that the results of two components  are significantly better than the performance of only one in each dimension, which reveals  that two components contribute to the flow of information and gradient transmission in the network.

\begin{figure}[t]
	\centering
	\includegraphics[height=5.2cm,width=0.38\textwidth]{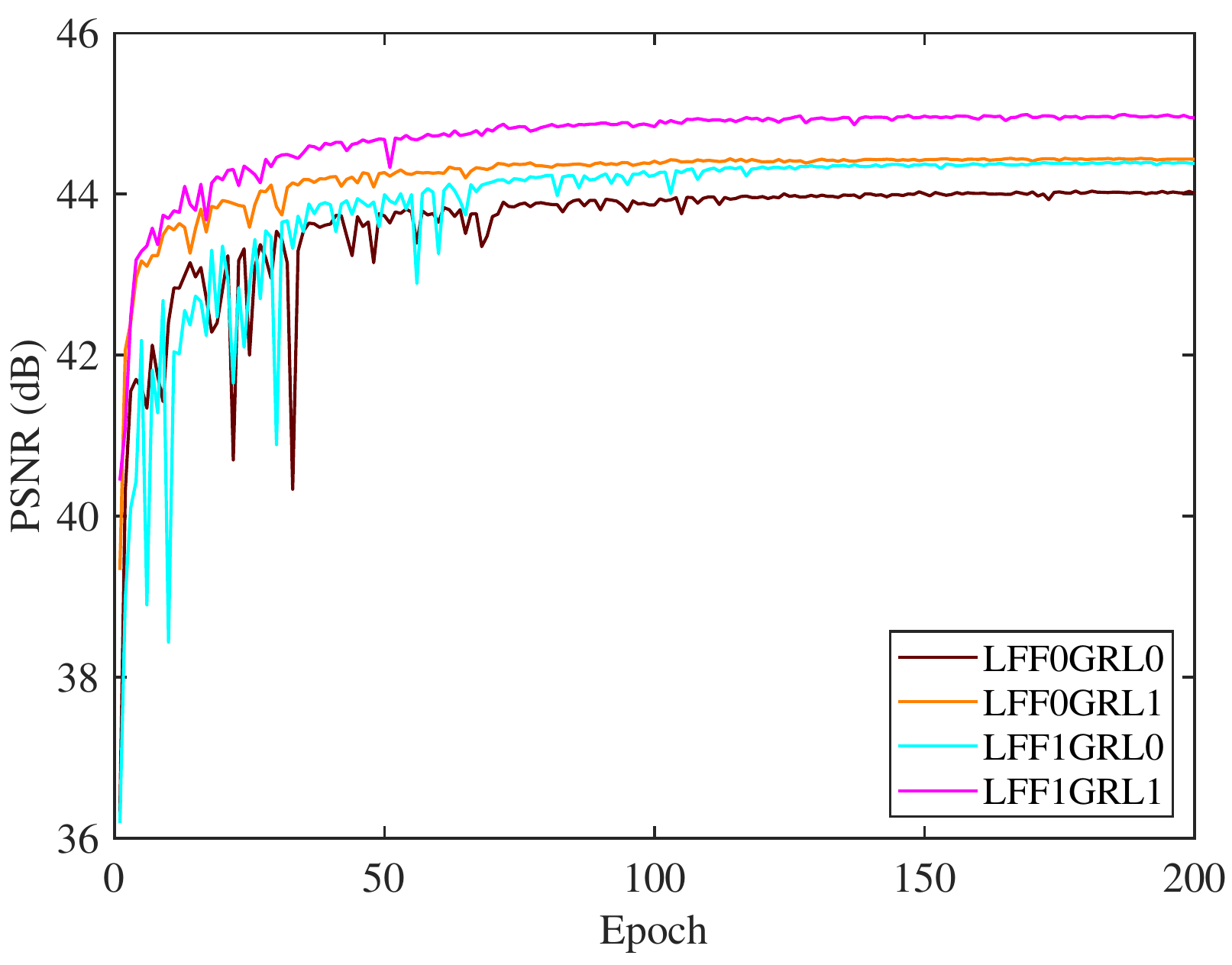}
	\caption{Ablation study of the the proposed method  for scale factor $\times$2 on CAVE dataset.}
	\label{fig:ablation_study}
\end{figure}

We also provide the convergence analysis only using PSNR for different combinations of components in Fig. \ref{fig:ablation_study}. One can observe that the convergence curve for LFF0GRL1 is more stable than that of LFF1GRL0 in the early iterations. Compared with baseline, LFF and GRL can effectively improve the performance in PSNR, which is consistent with the above analyses. To sum up, the analyses reveal that the effectiveness and benefits of the  proposed  LFF and GRL.

\begin{table}[tp]
	\centering
	\renewcommand\arraystretch{1.2}
	\caption{Ablation study about the components  for scale factor $\times$2 on CAVE dataset.}
	\label{table:ablation_study}
	\begin{tabular}{c|cccc}
		\toprule[0.8pt]	
		Components & \multicolumn{4}{c}{Different combinations of components} \\ \hline
		Local feature fusion (LFF) &$\times$ &\checkmark   &$\times$    &\checkmark    \\
		Global residual learning (GRL)  &$\times$ &$\times$    &\checkmark    &\checkmark   \\ \hline
		PSNR & 44.03  & 44.39  & 44.44 & 44.99 \\ 
		SSIM & 0.973  & 0.973  & 0.973 & 0.974 \\
		SAM  & 2.36   & 2.31   & 2.25  & 2.23 \\ \toprule[0.8pt]	
	\end{tabular}
\end{table}

\subsubsection{Study of Block}
\begin{table}[tp]
	\centering
	\renewcommand\arraystretch{1.2}
	\caption{Comparison of the performance of standard 3D convolution and separable 3D convolution.}
	\label{table:param}
	\begin{tabular}{c|cccc}
		\toprule[0.8pt]	
		Type & Parameters  &PSNR &SSIM  &SAM  \\ \hline
		Standard 3D convolution    & 2562K & 44.89 & 0.974 & 2.22             \\
		Separable 3D convolution   & 1275K & 44.99 &0.974 &2.23   \\ \toprule[0.8pt]	
	\end{tabular}
\end{table}
In this section, we study the efficiency of the proposed block using different types in module, including standard 3D convolution and separable 3D convolution.  The one is that we use block with separable 3D convolution, the other is standard 3D convolution that has removed ReLU activation function. Note that the convolution operations in initial feature extraction and image reconstruction are not replaced by separable 3D convolution in our network. The comparison results are shown in Table \ref{table:param}. Obviously, our proposed block can greatly reduce parameters (reduce ratio is 49.77\%), which  makes the network easier to train while reducing memory footprint. With respect to the results of PSNR,  using standard 3D convolution is lower than that of separable 3D convolution. We think that there are two main reasons for this problem:  1) there are too many parameters of the network, which makes the network more difficult to train;  2) the network all use the standard 3D convolution,  which makes the network pay too much attention to spectral information, so as to weaken the ability to learn spatial features. 

\subsubsection{Study of D}\label{ND}
The structure of our proposed SSRNet  is determined by the number of the spatial-spectral residual module $D$. To  analyze the effect of  parameters on the performance, we set  the range of $D$ from 2 to 5, and the results are displayed in Table. \ref{table:n}. One can observe that no matter what $D$ is, the values of SAM remain basically the same. Moreover, the values of PSNR and SSIM do not increase significantly when $D>3$. Although when $N$ is set to 5, the value of each evaluation index has been improved, especially for PSNR. However, it leads to a obvious increase in the corresponding network parameters.  Therefore,  we empirically set the parameter $D$ to 3 in our paper.

\begin{table}[]
	\centering
	\renewcommand\arraystretch{1.2}
	\caption{Analysis of the influence of the number of spatial-spectral residual modules $D$ on the performance.}
	\label{table:n}
	\begin{tabular}{c|cccc}
		\toprule[0.8pt]
		Evaluation Metrics & 2     & 3     & 4     & 5     \\ \hline
		PSNR  & 44.93 & 45.01 & 44.99 & 45.13 \\
		SSIM  & 0.971 & 0.974 & 0.974 & 0.975 \\
		SAM   & 2.24  & 2.23  & 2.23  & 2.22  \\ \toprule[0.8pt]
	\end{tabular}
\end{table}

\subsection{Comparisons with the State-of-the-art Methods}
\begin{table*}[tp]
	\centering
	\renewcommand\arraystretch{1.2}
	\caption{Quantitative evaluation of state-of-the-art SR algorithms by average PSNR/SSIM/SAM for different scale factors. The bold indicates the best performance.}
	\label{table:three}
	\begin{tabular}{c|c|ccccc}
		\toprule[0.8pt]
		\multirow{2}{*}{Dataset} & \multirow{2}{*}{Scale factor} & Bicubic & GDRRN & 3D-FCNN & EDSR & SSRNet \\
		&  & PSNR / SSIM / SAM  & PSNR / SSIM / SAM  & PSNR / SSIM / SAM  & PSNR / SSIM / SAM  & PSNR / SSIM / SAM  \\ \hline
		\multirow{3}{*}{CAVE} & $\times$2 & 40.76 / 0.962 / 2.66 & 41.66 / 0.965 / 3.84  & 43.15 / 0.968 / 2.30 & 43.86 / 0.973 / 2.63 &  \textbf{44.99} /   \textbf{0.974}  /  \textbf{2.26}\\
		& $\times$3 & 37.53 / 0.932 / 3.52 & 38.83 / 0.940 / 4.53 & 40.21 / 0.945 / 2.93 & 40.53 / 0.951 / 3.17 &   \textbf{40.90} /  \textbf{0.952} /  \textbf{2.81} \\
		& $\times$4 & 35.75 / 0.907 / 3.94 & 36.96 / 0.916 / 5.16 & 37.62 / 0.919 / 3.36 & 38.58 / 0.929 / 3.80 &   \textbf{38.94} /   \textbf{0.931} /  \textbf{3.29} \\ \hline
		\multirow{3}{*}{Harvard} & $\times$2 & 42.83 / 0.971 / 2.02 & 44.21 / 0.977 / 2.27 & 44.45 / 0.977 / 1.89 & 45.48 / 0.982 / 1.92 &   \textbf{46.25} /   \textbf{0.983} /   \textbf{1.88}\\
		& $\times$3 & 39.44 / 0.941 / 2.32 & 40.91 / 0.952 / 2.62 & 40.58 / 0.948 / 2.23 & 41.67 / 0.959 / 2.38 &  \textbf{42.65} /  \textbf{0.963} /  \textbf{2.20}\\
		& $\times$4 & 37.22 / 0.912 / 2.53 & 38.60 / 0.925 / 2.79 & 38.14 / 0.918 /  \textbf{2.36} & 39.17 / 0.932 / 2.56  &  \textbf{40.00} /  \textbf{0.937} / 2.41  \\ \hline
		\multirow{3}{*}{Foster} & $\times$2 & 55.15 / 0.998 / 4.39 & 53.52 / 0.996 / 5.63 &  \textbf{60.24} / 0.999 / 5.27 & 57.37 / 0.998 / 5.75 &  58.85 /   \textbf{0.999} /   \textbf{4.06}\\
		& $\times$3                         & 50.96 / 0.994 / 5.35 & 50.46 / 0.992 / 6.83 &  \textbf{55.55} / 0.996 / 6.30 & 52.98 / 0.996 / 7.71 & 54.93 /  \textbf{0.997} /  \textbf{5.13}\\
		& $\times$4                         & 48.28 / 0.988 / 5.99 & 47.83 / 0.987 / 7.69 & 52.18 / 0.992 / 7.79 & 50.36 / 0.992 / 7.10  &  \textbf{52.21} /  \textbf{0.994} /  \textbf{5.70}  \\ 		\toprule[0.8pt]
	\end{tabular}
\end{table*}

\begin{figure*}[tp]
	\centering
	\includegraphics[height=20cm, width=1\textwidth]{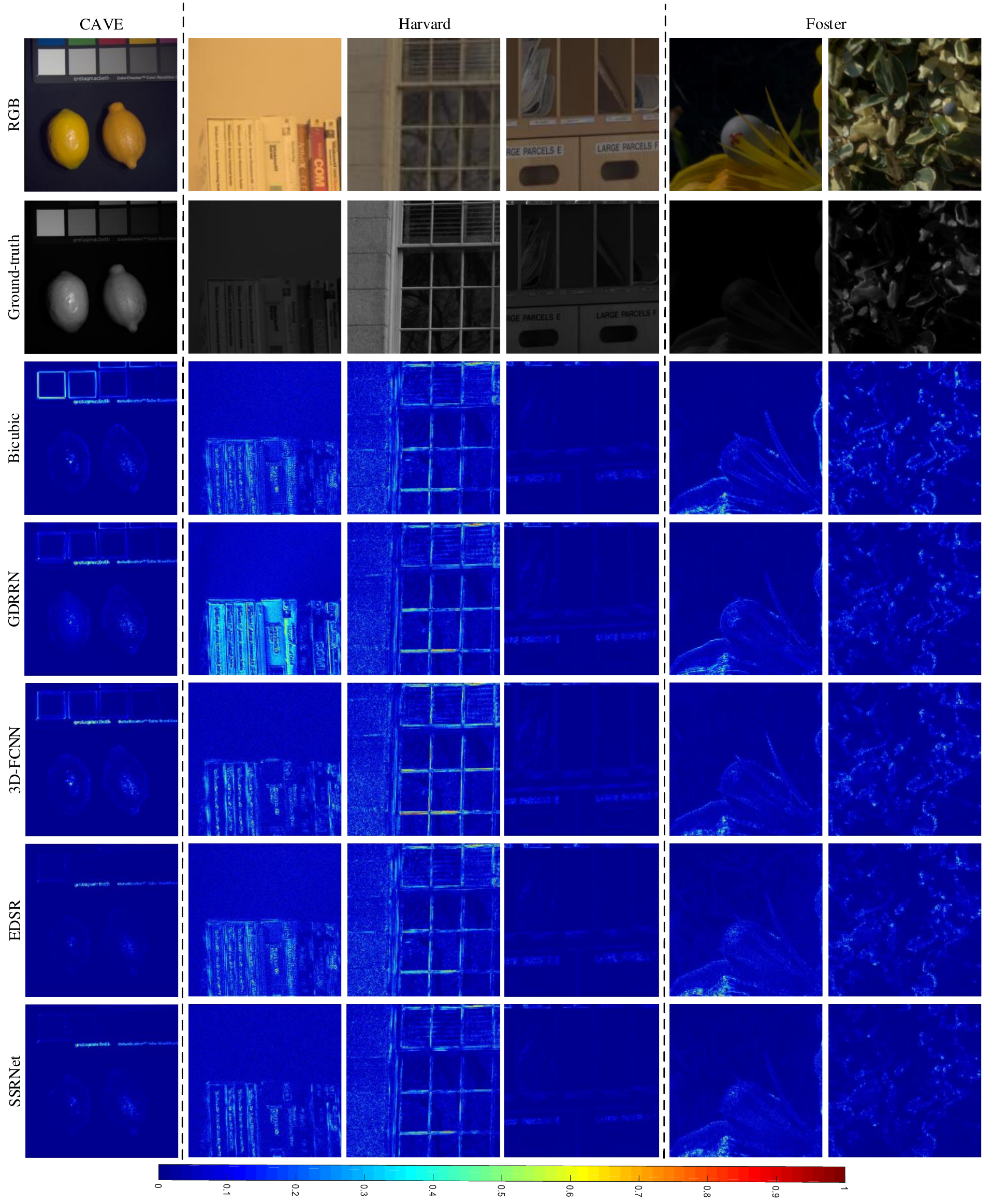}
	\caption{Absolute error map comparisons of our SSRNet  with existing methods for scale factor $\times$4.}
	\label{fig:visual_2}
\end{figure*}

\begin{figure*}[t]
	\centering
	\subfigure[]{
		\includegraphics[width=0.32\textwidth]{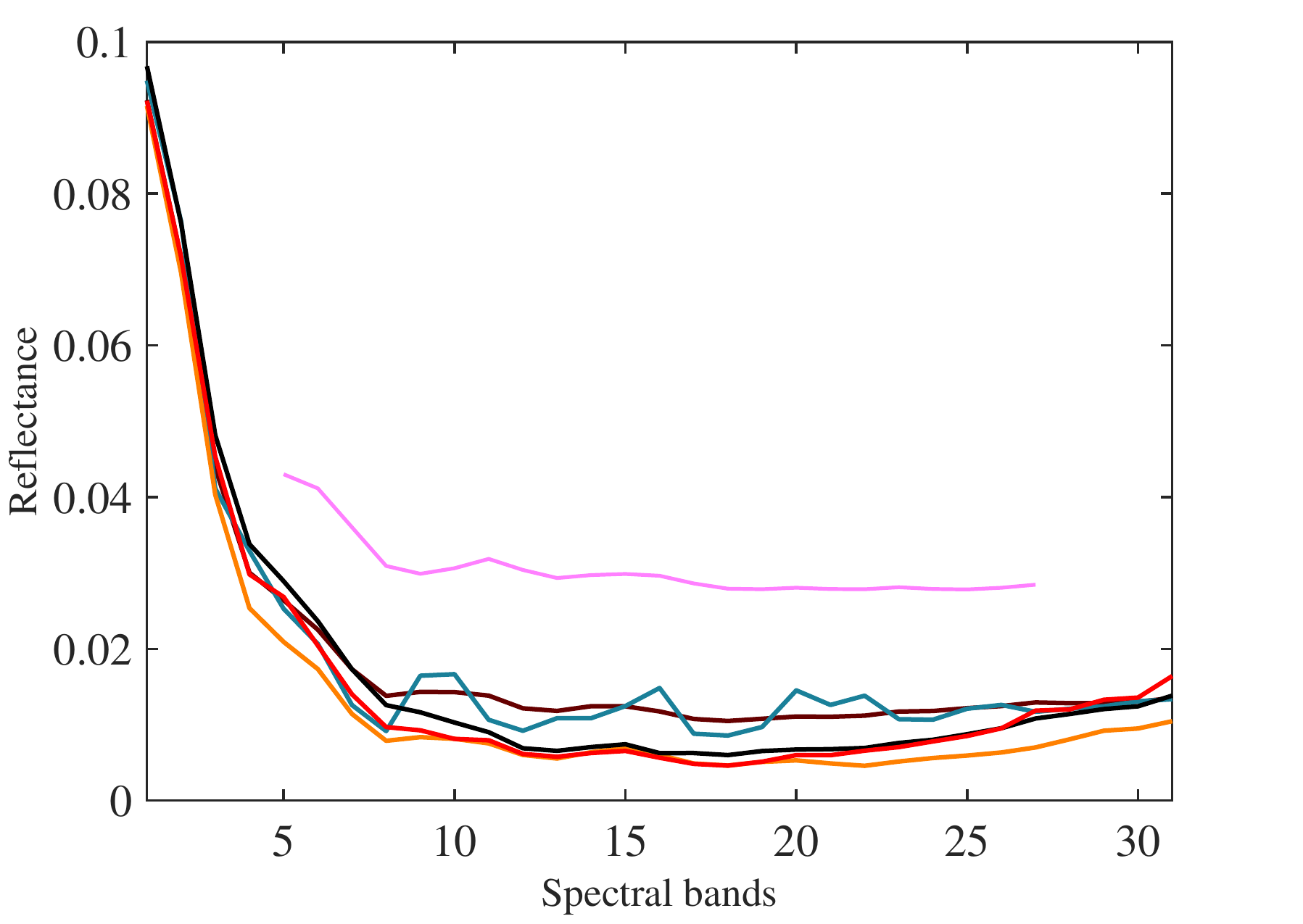}}
	\subfigure[]{
		\includegraphics[height=4.15cm, width=0.32\textwidth]{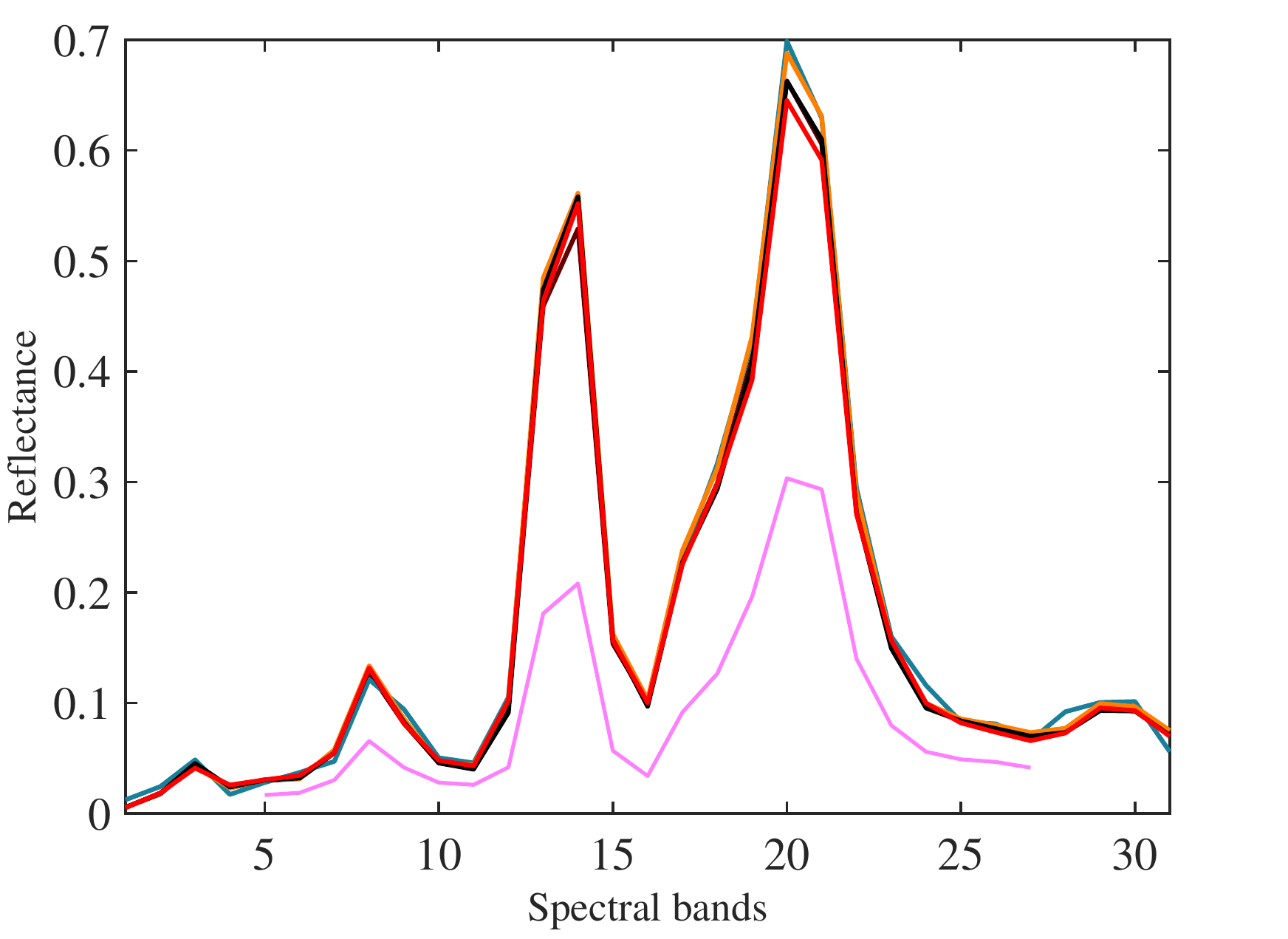}}
	\subfigure[]{
		\includegraphics[width=0.32\textwidth]{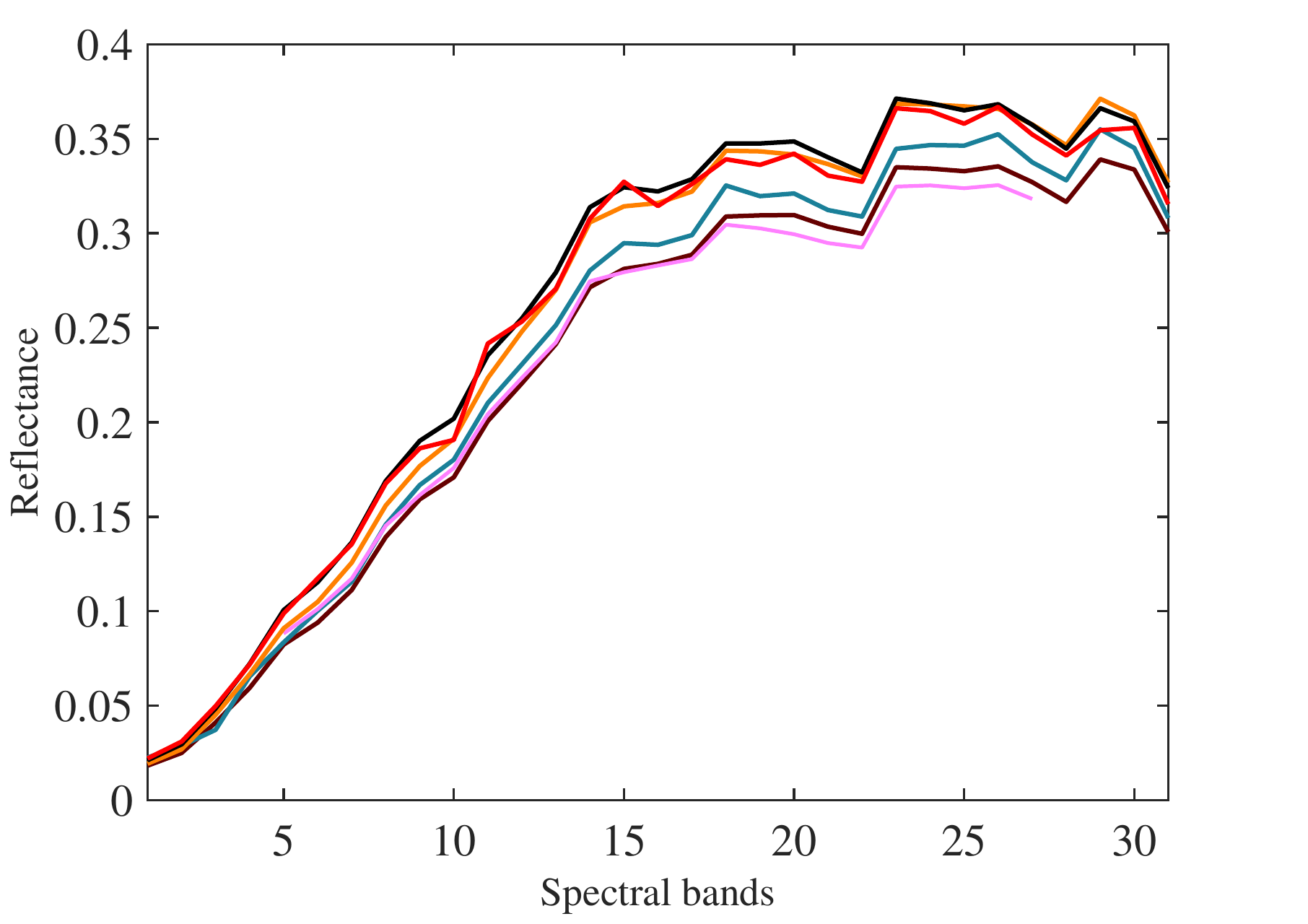}}
	
	\subfigure[]{
		\includegraphics[width=0.32\textwidth]{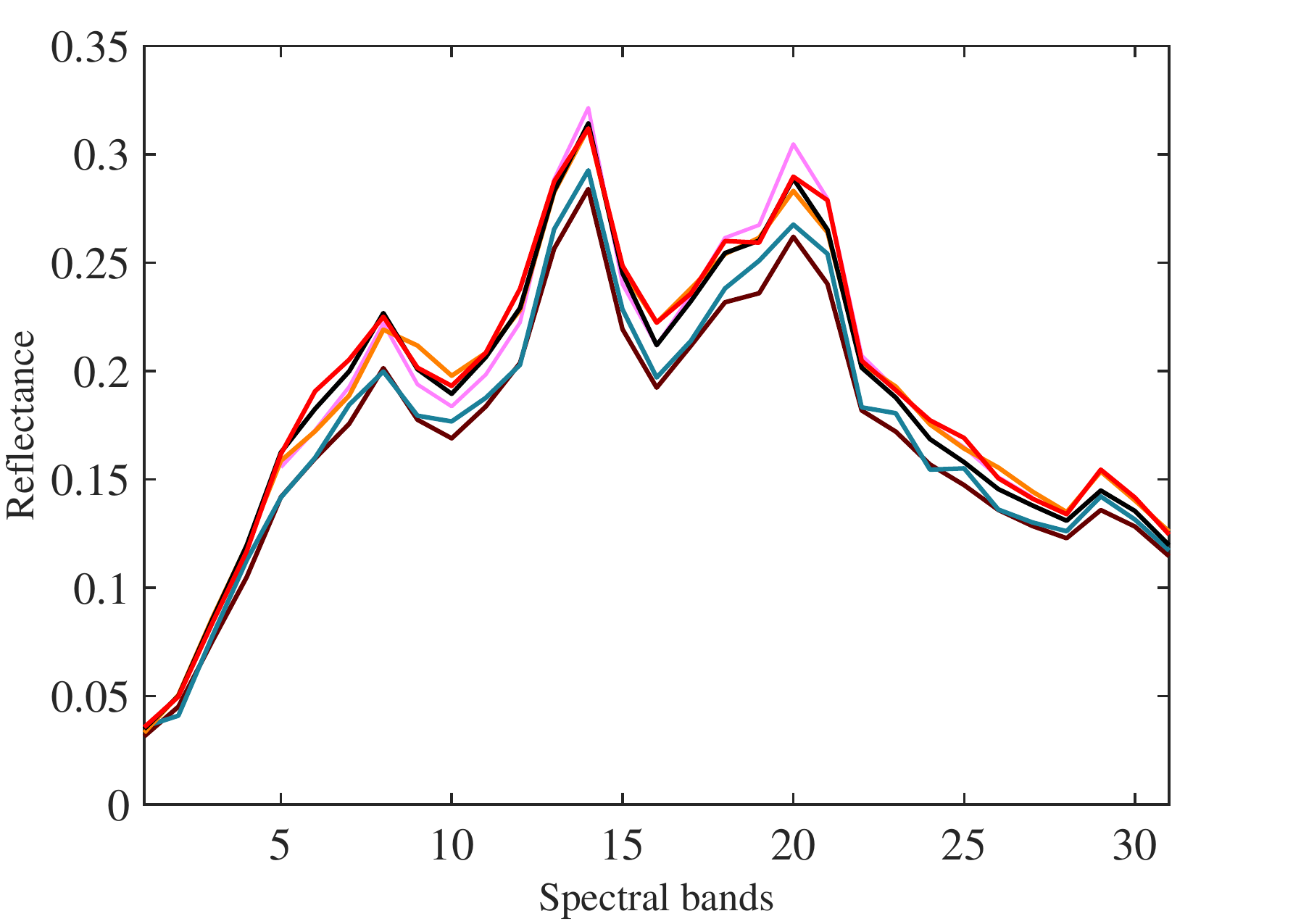}}	
	\subfigure[]{
		\includegraphics[width=0.32\textwidth]{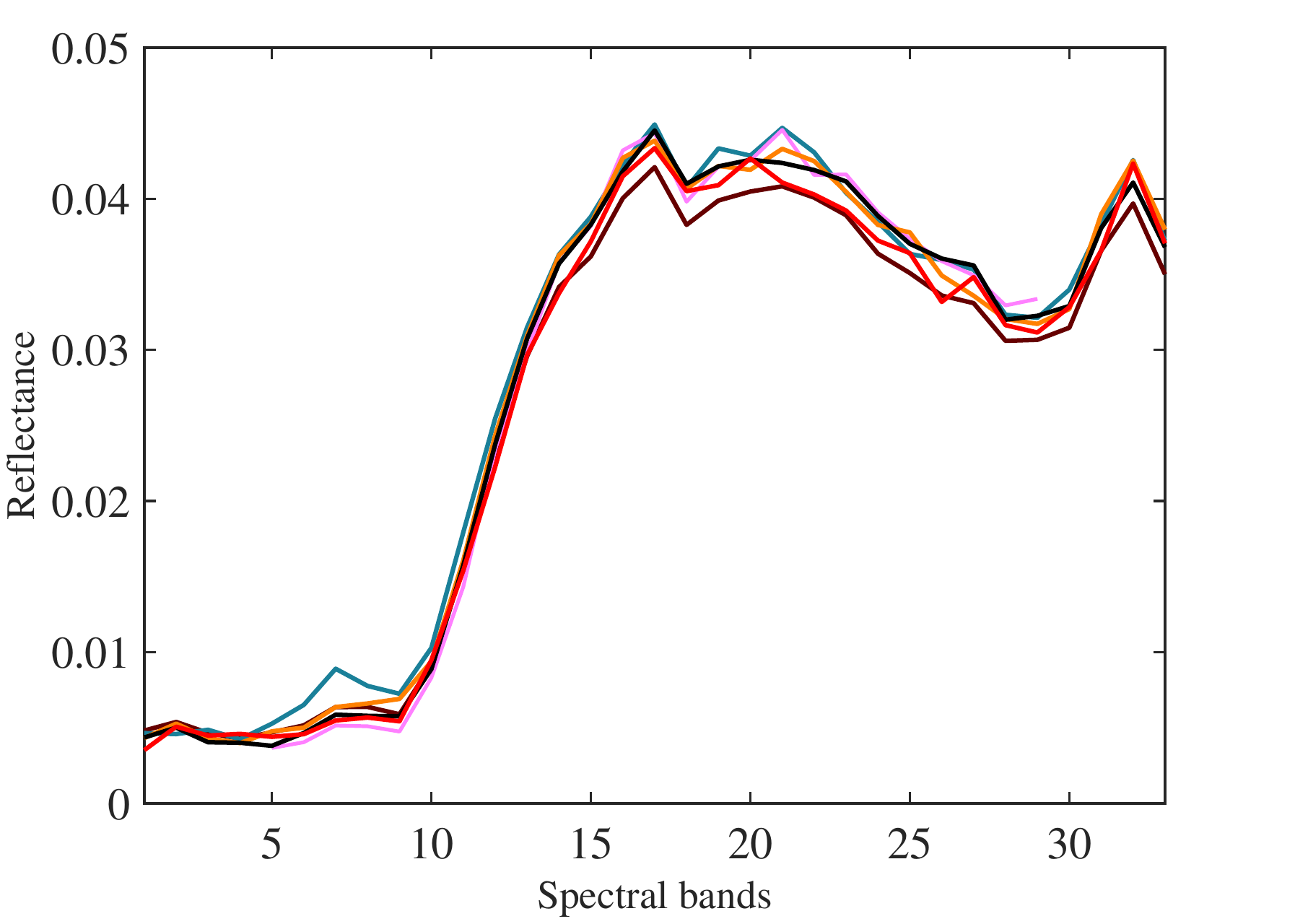}}
	\subfigure[]{
		\includegraphics[width=0.32\textwidth]{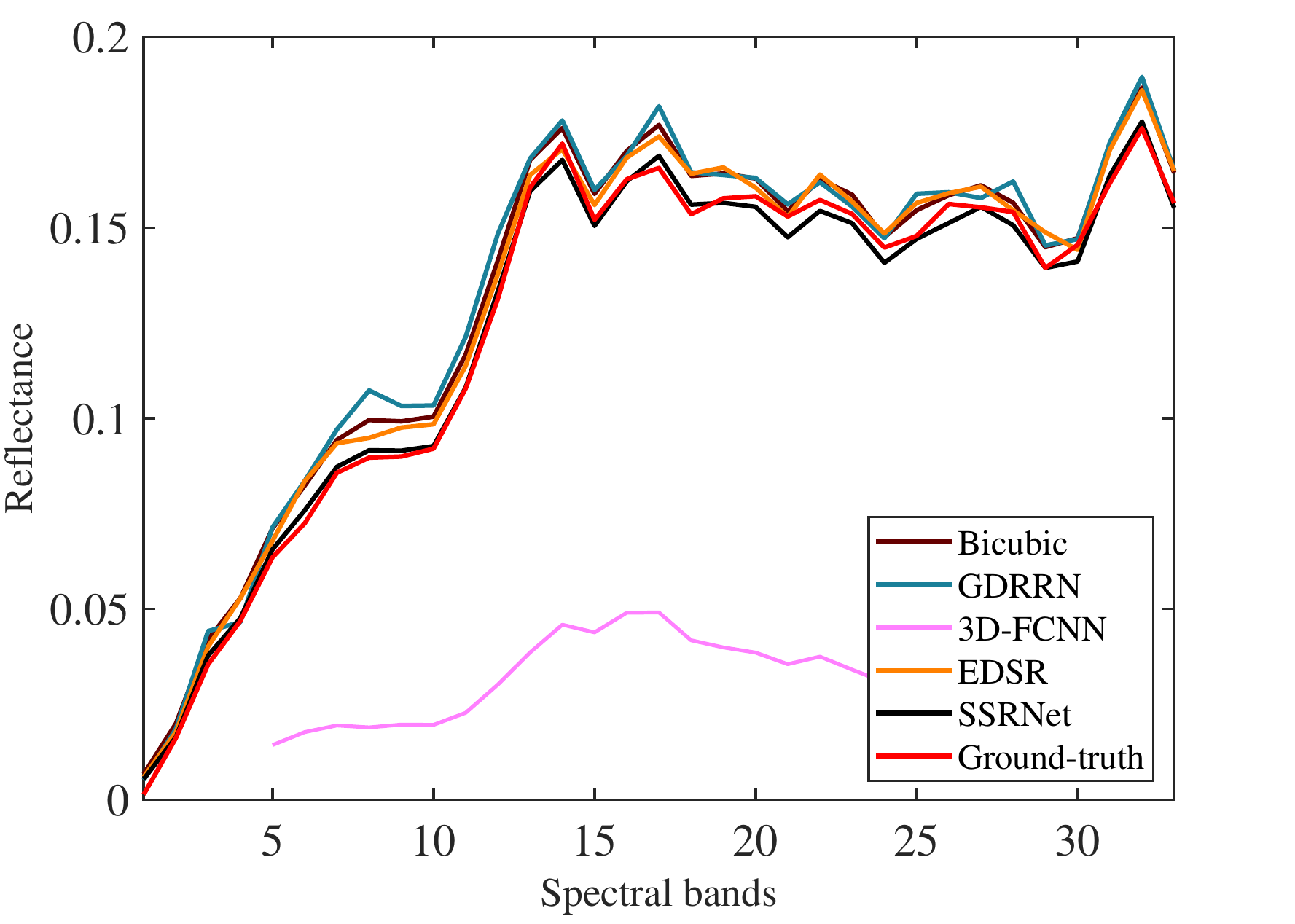}}					
	\caption{Spectral  distortion  comparisons by randomly selecting a pixel. (a)-(f)  Results of the spectral curves of six scenes, respectively.}
	\label{fig:spectral}
\end{figure*}
In this section, we adopt three public hyperspectral image datasets to evaluate the effectiveness of our SSRNet with existing SR approaches using three evaluation indexes. Table \ref{table:three} depicts the  quantitative evaluation of state-of-the-art SR algorithms by average PSNR/SSIM/SAM for different scale factors.

As shown in table, our method can achieve the best results  than other algorithms on CAVE dataset. Specifically, the Bicubic  produces  the worst performance  among these competitors. For the GDRRN algorithm, all the results are slightly higher than the worst Bicubic but  lower than other methods. It is caused by the addition of a SAM item in the loss function. As a result, the network can not optimize the difference between  reconstructed  and  high-resolution images. Furthermore, the results of 3D-FCNN in PSNR and SSIM are lower than that of EDSR, but the performance in SAM of 3D-FCNN is obviously higher than that of EDSR, which is due to the fact that 3D-FCNN uses 3D convolution to extract the  spectral features of hyperspectral image. Thus, this algorithm can  void the spectral distortion of the reconstructed hyperspectral image well. However,  the image  obtained by 3D-FCNN lose part of the bands (the algorithm only obtains 23 bands on hyperspectral image with 31 bands), which is not suitable for image SR.  Compared with the existing SR approaches, our method obtains excellence performance. The proposed method is significantly superior to the scale factor $\times 4$ of the second performance algorithm (EDSR) in terms of three evaluation metrics (+0.36dB, +0.002, and -0.07). 

Similarly, except for 3D-FCNN,  the SSRNet outperforms other competitors in three aspects on  Hararvd dataset. Concretely, unlike on CAVE dataset, GDRRN and 3D-FCNN has achieved approximately the same results, because the number of hyperspectral images on augmented Harvard dataset is more than that  on CAVE dataset. This is more beneficial to network training with many parameters, such as EDSR. Moreover, it also enables our approach to achieve higher performance (+0.98dB, +0.004, and -0.03) on this dataset than on CAVE dataset for scale factor $\times 3$.  Likewise, the proposed approach achieves  good  performance in comparison to existing state-of-the-art methods on Foster dataset, particularly  in SSIM and SAM.

In Fig. \ref{fig:visual_2}, we show visual comparisons with different algorithms for scale factor $\times 4$  on three datasets. The figure only provides visual results of the $27$-th band of six typical scenes. As revealed in the figure, the ground-truth is grey. So in order to observe the difference between reconstructed hyperspectral image and ground-truth clearly, the absolute error map between them is presented. In general, the bluer the absolute error map is, the better the restored image is. Note that each hyperspectral image is normalized. From this figure, we can see that our proposed SSRNet obtains very low absolute error results. In some regions, especially for the edges of the image, our method generates shallow edge information with little or no edge information. It means our proposed SSRNet  generates more realistic visual results compared with other methods, which is consistent with our analysis in Table \ref{table:three}.

We also visualize the spectral distortion of the reconstructed image by drawing spectral curves for six scenes, which is presented in Fig. \ref{fig:spectral}. Since 3D-FCNN loses some of the bands during reconstruction, we only show some of bands. It can be seen from this figure that the distortion for 3D-FCNN is the most severe. The distortion of the spectral curve obtained by Bicubic is relatively small compared with 3D-FCNN. Moreover, among these competitors, the spectral curves of GDRRN, EDSR, and SSRNet are basically consistent with that of ground-truth, but  the results of our method  are much closer to the ground truth in most cases, which proves our algorithm  attains higher spectral fidelity. In conclusion,  SSRNet can not only outperform state-of-the-art SR algorithms through  quantitative evaluation, but also  yield more realistic visual results. 
 
\section{Conclusion}
Considering that existing deep learning-based hyperspectral image super-resolution (SR) methods can not simultaneously explore spatial information and spectral information between bands, we develop a novel spectral-spatial residual network (SSRNet) to reconstruct hyperspectral image, claiming the following contributions: 1) without changing the size of the hyperspectral image, our proposed network adopts 3D convolution to effectively exploit spatial and spectral features instead of 2D convolution; 2) we propose spatial-spectral residual module (SSRM). The module can make full use of the hierarchical features generated by each unit and learn effective features adaptively by the way of local feature fusion; and 3) we employ separable 3D convolution  to extract spatial and spectral features respectively,  which reduces the training parameters of the network, thus making the network easier to train. Extensive benchmark evaluations well demonstrate that  our SSRNet can not only outperform state-of-the-art SR algorithms, but also yield more realistic visual results. 

In the feature, we plan to  improve proposed SSRNet by two aspects. The one is to increase the spatial feature extraction capability of the network and reduce the feature extraction between spectra in unit. Second, hybrid 3D/2D convolution is adopted to reduce the training complexity of each module and thus accelerate the execution speed of the network.

\bibliographystyle{IEEEbib}

\bibliography{refs}	
\end{document}